\pdfoutput=1

\documentclass[11pt]{article}

\usepackage{acl}

\usepackage{times}
\usepackage{latexsym}
\usepackage{hyperref}
\usepackage{url}
\usepackage{times}
\usepackage{latexsym}
\usepackage{booktabs}
\usepackage{graphicx}
\usepackage{subcaption}
\usepackage{multirow}
\usepackage{arydshln}
\usepackage{wrapfig}
\usepackage{algorithm}
\usepackage{algpseudocode}
\usepackage{amssymb}

\usepackage[T1]{fontenc}

\usepackage[utf8]{inputenc}

\usepackage{microtype}

\usepackage{inconsolata}

\title{DynaThink: Fast or Slow? A Dynamic Decision-Making Framework for Large Language Models}

\author{Jiabao Pan  \\ Zhejiang University \\  \texttt{3200102835@zju.edu.cn} \And
        Yan Zhang \\ National University \\ of Singapore \\ \texttt{eleyanz@nus.edu.sg}   \And
        Chen Zhang \\ National University \\ of Singapore \\ \texttt{chen\_zhang@u.nus.edu}
        \AND
        Zuozhu Liu \\ Zhejiang University \\ zuozhuliu@intl.zju.edu.cn  \And
        Hongwei Wang \\ Zhejiang University \\ hongweiwang@intl.zju.edu.cn  \And
        Haizhou Li \\ The Chinese University of Hong Kong \\ haizhou.li@nus.edu.sg }

\begin{document}
\maketitle

\begin{abstract}
Large language models (LLMs) have demonstrated emergent capabilities across diverse reasoning tasks via popular Chains-of-Thought (COT) prompting. However, such a simple and fast COT approach often encounters limitations in dealing with complicated problems, while a thorough method, which considers multiple reasoning pathways and verifies each step carefully, results in slower inference. This paper addresses the challenge of enabling LLMs to autonomously select between fast and slow inference methods, thereby optimizing both efficiency and effectiveness. We introduce a dynamic decision-making framework that categorizes tasks into two distinct pathways: 'Fast', designated for tasks where the LLM quickly identifies a high-confidence solution, and 'Slow', allocated for tasks that the LLM perceives as complex and for which it has low confidence in immediate solutions as well as requiring more reasoning paths to verify.   Experiments on five popular reasoning benchmarks demonstrated the superiority of the DynaThink over baselines. 


\end{abstract}

\section{Introduction}
LLMs~\citep{openaichatgpt,Touvron2023Llama2O,OpenAI2023GPT4TR} have emerged as prominent foundation models for diverse applications due to their outstanding ability to understand and generate human-like text. One notable attribute of LLMs is their capability of COT reasoning~\citep{wei2022chain}, where they can perform multi-step reasoning using only a few demonstrations. However, the performance of such simple and fast COT prompting is not satisfactory for complex reasoning problems, which often need more careful thought and analysis~\citep{Stanovich2000IndividualDI}. To improve this, ~\citet{wang2022self} introduced a self-consistency strategy as a replacement for the previous straightforward decoding method used in COT prompting. This self-consistency strategy involves trying out multiple reasoning paths and selecting the most consistent answers, significantly improving LLM performance in different reasoning tasks. Recent advanced works ~\citep{Yao2023TreeOT,Besta2023GraphOT,Shinn2023ReflexionAA,Madaan2023SelfRefineIR,Miao2023SelfCheckUL}  expanded on this strategy by considering different reasoning paths and self-evaluation to make holistic decisions.  However, it's important to note that these methods require more resources and therefore slow down the decision-making process, especially for high-cost LLMs such as GPT-4 ~\citep{OpenAI2023GPT4TR}. This highlights the need for a method that lets LLMs adjust resource use based on the specific needs and complexities of tasks, which is crucial for developing and using LLMs effectively in real-world situations, allowing these models to provide the best value in various applications while managing resources wisely.

 \renewcommand{\dblfloatpagefraction}{.9}
\begin{figure*}[ht]
    \centering
    \includegraphics[scale=0.13]{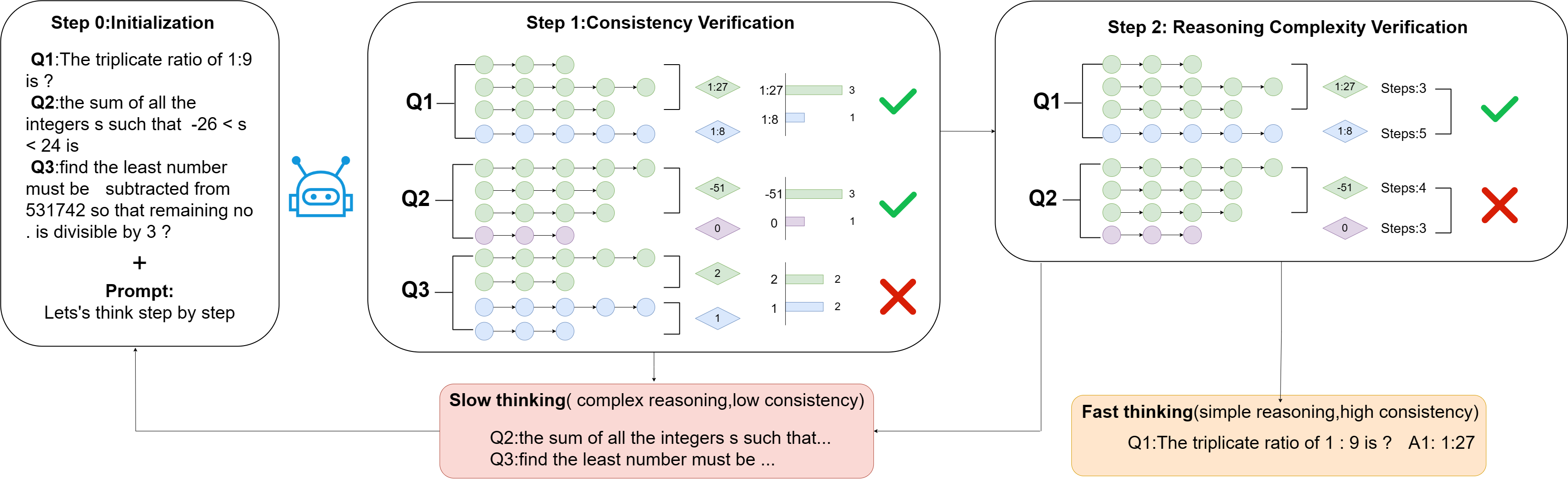}
    \vspace{-1mm}
    \caption{ The workflow starts by incorporating the widely used CoT prompt, like 'let's think step by step'  and then initially querying LLMs four times for each question (which can be fewer than four to begin with). In the first step, we choose question 1 and question 2 because they have one answer with more than half the votes. However, question 3 has a tie, with two answers getting equal votes, so we consider it a slow-thinking question. Next, we look at questions 1 and 2 to see how many steps are needed for each answer. For question 2, the answer with the most votes takes four steps, while the other answer needs three steps, so question 2 is also categorized as a slow-thinking question. Regarding question 1, we classify it as a fast-thinking question because the answer with the most votes also requires the fewest steps, allowing us to output this answer directly. However, for slow-thinking questions, we require additional iterations to make a selection. }
    \vspace{-3mm}
    \label{fig:framework}
\end{figure*}
To address the challenge, we introduce a dynamic decision-making framework called "DynaThink." This framework helps LLMs make smart choices between quick and thorough ways of solving problems. It optimizes a balance between being efficient and effective in various reasoning tasks. In this framework, we categorize tasks into two groups: "Fast" for tasks where LLMs can quickly find confident answers, and "Slow" for more complex tasks that require thorough exploration of different reasoning paths to confirm answers.  Establishing rules to distinguish between fast and slow reasoning is challenging. We adopt two criteria including consistency verification and reasoning complexity verification. The principle of consistency verification is reflective of the human heuristic, wherein the concurrence of multiple distinct thought processes on a singular answer augments the confidence of its correctness~\citep{Kahneman2011ThinkingFA}.
Different from the straightforward majority voting used in ~\citep{wang2022self}, which sometimes leads to problems like tie-breakers, we only validate answers when they secure more than half of the total votes, indicating they are high-confidence solutions with the best votes.  On the other hand, our reasoning complexity verification criteria are based on reasoning processes with a minimum number of reasoning steps. LLMs can make mistakes during the reasoning process~\citep{Madaan2023SelfRefineIR, Ling2023DeductiveVO}, while each reasoning step of LLMs can be thought of as a decision made under uncertainty. The accumulation of these decisions can lead to a compounding of uncertainty, reducing the overall confidence in the final decision.  Therefore, fewer reasoning steps often yield more reliable and confident outcomes due to reduced error propagation.To empirically corroborate the two criteria, we have conducted experiments on multiple reasoning tasks and found a strong correlation between the efficacy of LLMs in problem-solving and both the length of their reasoning paths and the voting of the outputs. Questions that meet these criteria are sorted using simple and fast decision making processes.  The rest of the questions are subject to more exploration of different reasoning paths and comprehensive self-evaluations~\citep{Yao2023TreeOT,Besta2023GraphOT,Miao2023SelfCheckUL}. 

\section{DynaThink}
DynaThink, as shown in Figure~\ref{fig:framework} \footnote{The algorithm presudo code is shown in appendix ~\ref{sec；algorithm}}, starts by setting the initial query times for LLM, which in this case is 2. The LLM is then queried with the problem set $P$, yielding the response set $Q$. First, we employ consistency verification to analyze the diversity of answers each question receives. Questions whose answers get over half the votes are placed in $Q_1$, serving as potential fast-thinking questions. The rest go into $Q_2$ for more review in the following rounds with added resources. Next, we utilize reasoning complexity verification to examine the step counts needed for each response, picking the one with the fewest steps throughout all generations. We then see if this selected answer from $Q_1$ indeed has the least steps. If it does, it goes into $Q_3$ as a fast-thinking question of this round. If not, it returns to $Q_2$ for more rounds of checking. Lastly, we see if we've picked any fast-thinking questions this round. If $Q_3$ has questions, we proceed to another iteration; if it's empty, we conclude, leaving the questions in $Q_2$ as the slow-thinking ones. For the fast-thinking questions, we can easily arrive at an answer (more than half vote + least number of reasoning steps). For the slow questions, we repeat the same CoT + self-consistency procedure with more queries to the LLMs so as to identify the answer with more than half vote and the least number of reasoning steps.

\begin{figure*}[htbp]
    \centering
    \begin{subfigure}{0.3\textwidth}
        \centering
        \includegraphics[width=\linewidth]{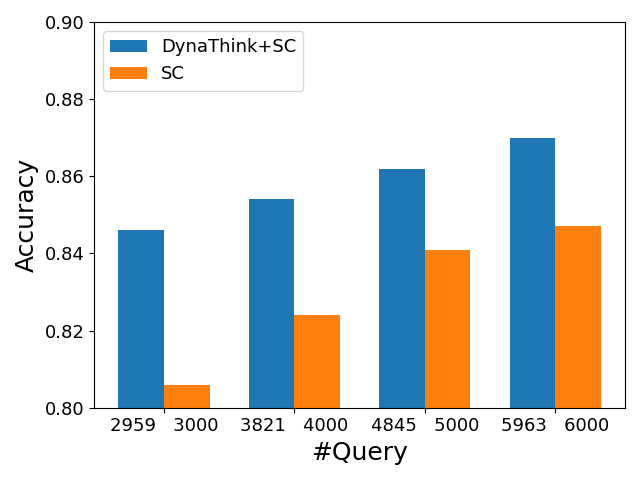}
        \caption{SVAMP(zero-shot-GPT-3.5-turbo)}
        \label{fig:sub1}
    \end{subfigure}
    \hfill
    \begin{subfigure}{0.3\textwidth}
        \centering
        \includegraphics[width=\linewidth]{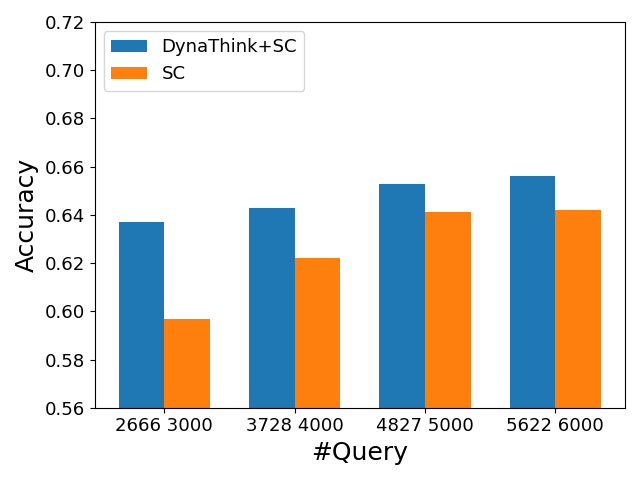}
        \caption{SQA(zero-shot-GPT-3.5-turbo)}
        \label{fig:sub2}
    \end{subfigure}
    \hfill
    \begin{subfigure}{0.3\textwidth}
        \centering
        \includegraphics[width=\linewidth]{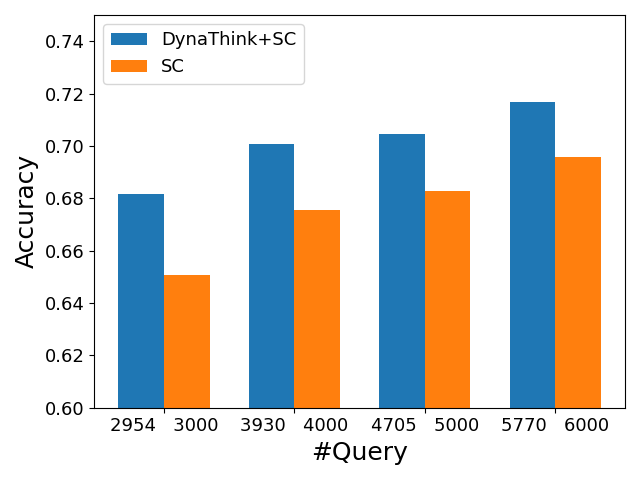}
        \caption{MathQA(zero-shot-GPT-3.5-turbo)}
        \label{fig:sub2}
    \end{subfigure}
    \begin{subfigure}{0.3\textwidth}
        \centering
        \includegraphics[width=\linewidth]{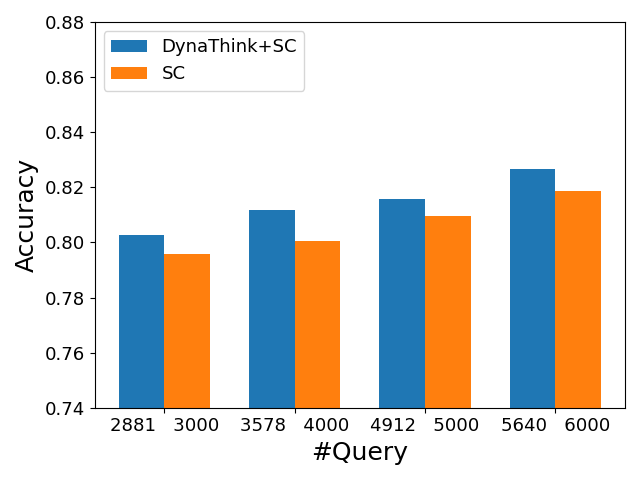}
        \caption{GSM8K(zero-shot-Gemini)}
        \label{fig:sub2}
    \end{subfigure}
    \hfill
    \begin{subfigure}{0.3\textwidth}
        \centering
        \includegraphics[width=\linewidth]{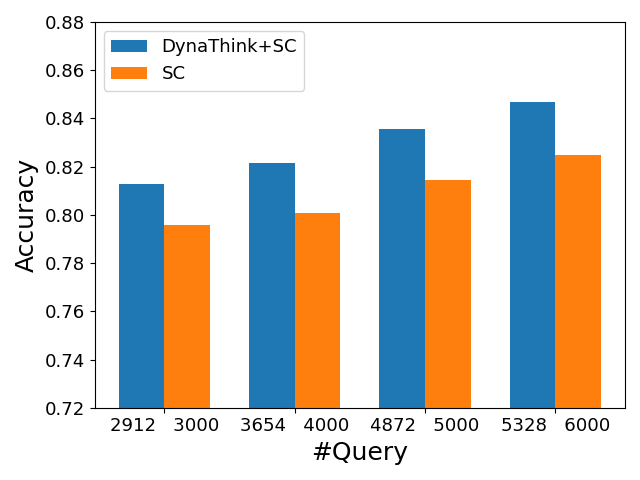}
        \caption{GSM8K(zero-shot-GPT-4)}
        \label{fig:sub2}
    \end{subfigure}
    \hfill
    \begin{subfigure}{0.3\textwidth}
        \centering
        \includegraphics[width=\linewidth]{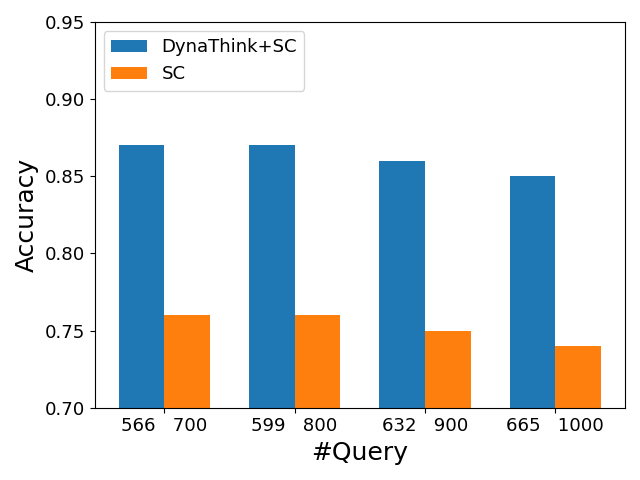}
        \caption{GSM8K(Mixtral-8x7B-v0.1)}
        \label{fig:sub1}
    \end{subfigure}
    
    \caption{ SC: the original self-consistency approach that uses majority voting to identify the most agreed-upon answer~\citep{wang2022self}. DynaThink+SC:  divides the question sets into fast and slow-thinking categories and applies different selection criteria to the answers from each set. The CoT prompting technique is utilized by both SC and DynaThink+SC, and this is applied in both zero-shot and few-shot settings ~\citep{wei2022chain,Kojima2022LargeLM}.  To ensure a fair comparison, DynaThink+SC also utilizes the SC strategy for the slow-thinking question set, to determine the final answer. However, there is a slight adjustment in the number of queries for LLM to accommodate the different processing requirements of the fast and slow-thinking question sets. It's important to highlight that we always ensure that the operational cost of using DynaThink+SC is either lower or competitive with the use of the SC strategy. In essence, the goal of DynaThink+SC is to optimize efficiency and cost-effectiveness.  SQA means StrategyQA. Due to space limitations, more results are presented in appendix~\ref{sec:The rest results of main result}. }
    \label{fig:main}
\end{figure*}


\section{Experiments}





\paragraph{Setup} We evaluate the efficiency and effectiveness of  DynaThink  on six diverse reasoning datasets, i.e., StrategyQA~\citep{Geva2021DidAU}, GSM8K~\citep{cobbe2021gsm8k}, MATH~\citep{Hendrycks2021MeasuringMP}, AQUA-RAT~\citep{Ling2017ProgramIB}, SVAMP~\citep{Patel2021AreNM} and MATHQA~\citep{Amini2019MathQATI}.\footnote{The statistics can be found in the Appendix.}  We use three popular black-box LLMs like GPT-3.5-turbo, GPT-4, Gemini as well as one popular open-source LLM, i.e., Mixtral-8x7B~\citep{Jiang2024MixtralOE} as the backbones. We employ the self-consistency strategy~\citep{wang2022self} within the CoT reasoning as our main baseline following~\citep{Ling2023DeductiveVO,Lightman2023LetsVS, Miao2023SelfCheckUL}. Although it's simple, it's still a strong method according to current research.

\renewcommand{\dblfloatpagefraction}{.9}
\begin{figure*}[h]
    \centering
    \begin{subfigure}{0.3\textwidth}
        \centering
        \includegraphics[width=\linewidth]{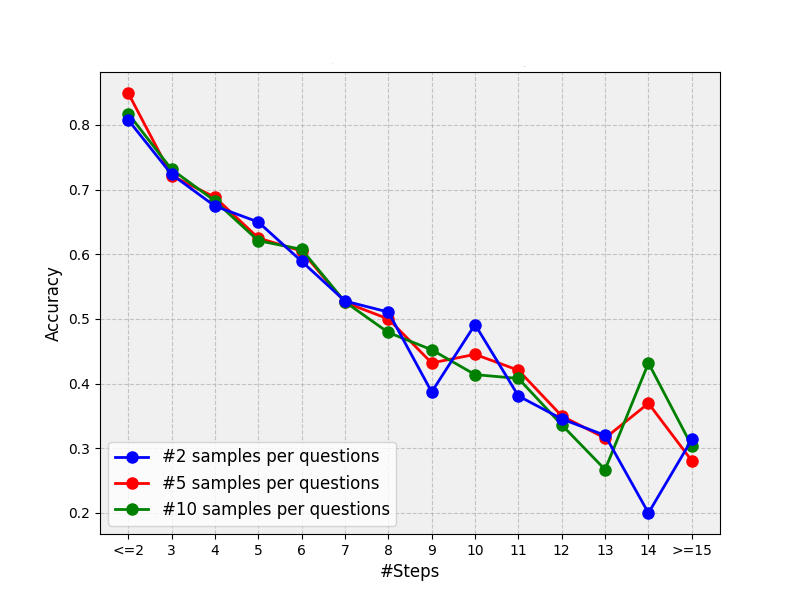}
        \caption{AQuA(zero shot)}
        \label{fig:sub1}
    \end{subfigure}
    \hfill
    \begin{subfigure}{0.3\textwidth}
        \centering
        \includegraphics[width=\linewidth]{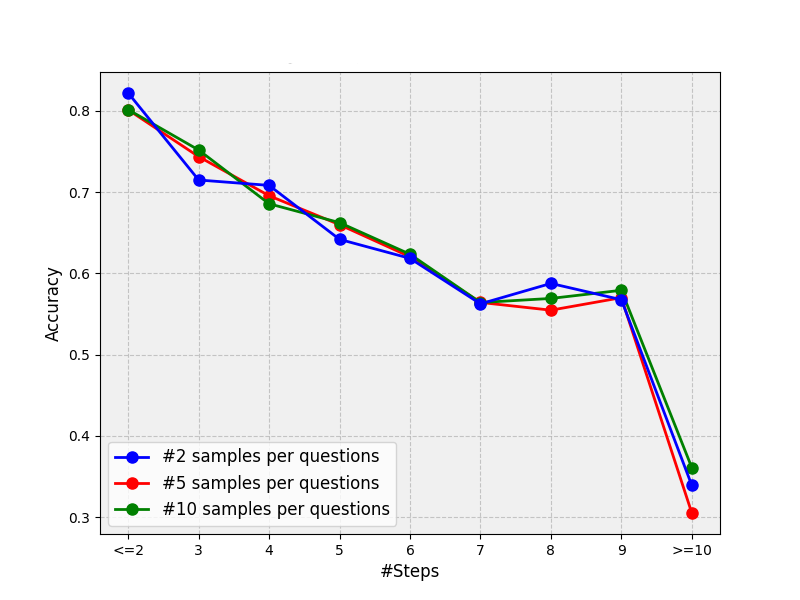}
        \caption{GSM8K(zero shot)}
        \label{fig:sub2}
    \end{subfigure}
    \hfill
    \begin{subfigure}{0.3\textwidth}
        \centering
        \includegraphics[width=\linewidth]{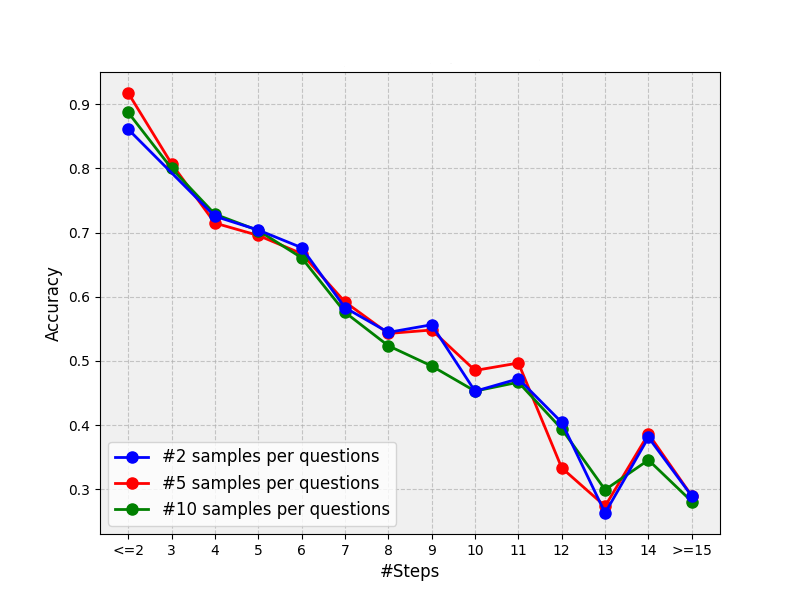}
        \caption{MathQA(zero shot)}
        \label{fig:sub1}
    \end{subfigure}

    \caption{Correlation between the distribution of reasoning steps and reasoning performance. We employed the self-consistency strategy within the zero-shot and few-shot CoT prompting techniques, to query the GPT-3.5-Turbo model two, five, and ten times for each question.Due to space limitations, more results are presented in appendix~\ref{sec:rest data of ablation study of reasoning complexity verification}}
    \label{fig:reasoning-complexity}
        \vspace{-3mm}
\end{figure*}
\paragraph{Main Results}
Figure~\ref{fig:main} presents the main results. We can observe that DynaThink consistently enhances both the effectiveness and efficiency of SC across a range of reasoning tasks in both zero-shot and few-shot scenarios. For instance, within the zero-shot MATH setting, utilizing GPT-3.5-Turbo, DynaThink attains an accuracy rate of 45\% with 2758 LLM queries, outperforming SC, which secures an accuracy of 41.9\% using the same number of LLM queries. Likewise, in the few-shot MathQA context, our approach exhibits a 4\% improvement in accuracy with 2849 LLM queries, in contrast to SC's performance with 3000 LLM queries. The integration of consistency checks and reasoning complexity verification as criteria is deemed crucial for this enhanced performance. These criteria provide an advanced method for accurately evaluating confidence levels, minimizing the build-up of uncertainties, and improving resource allocation and efficiency, thereby significantly increasing the accuracy of solutions. 

In addition, we can find that, GPT-4 demonstrates the best performance among the three evaluated LLMs. Despite the high baseline established by such a top-performing LLM, DynaThink is still able to exhibit superior performance. For example, in the zero-shot MATHQA scenario, DynaThink achieves an accuracy of 73.8\% with 2827 LLM queries, surpassing SC, which attains an accuracy of 71.7\% with 3000 LLM queries. Similarly, in the zero-shot GSM8K scenario, DynaThink achieves an accuracy of 81.6\% with 2912 LLM queries, exceeding SC's performance, which is 79.4\% accuracy with 3000 LLM queries. These findings suggest that the DynaThink framework generalizes effectively across various LLMs.

\paragraph{Ablation Study of Reasoning Complexity Verification}
We study the correlation between the distribution of reasoning steps and reasoning performance. Specifically, we conducted a comprehensive analysis by randomly selecting 200 questions from each of the AQuA, GSM8K, and MathQA datasets to compile a development set. The study employed the self-consistency strategy, within the zero-shot and few-shot CoT prompting techniques to query the GPT-3.5-Turbo model two, five, and ten times per question. The findings, illustrated in Figure~\ref{fig:reasoning-complexity}, indicate a consistent pattern: an increase in the number of reasoning steps correlates with a reduction in the LLM’s reasoning performance. This trend is evident across both zero-shot and few-shot learning paradigms.

The results from this investigation lend substantial support to our initial hypothesis, which posits that minimizing the number of reasoning steps serves as a practical metric for verifying reasoning complexity. This approach is pivotal in assessing the confidence level of the LLMs' reasoning outcomes, offering valuable insights into achieving an ideal equilibrium between the extent of reasoning steps and the efficacy of model performance. It underscores the significance of maintaining efficiency and succinctness in the reasoning process to procure responses from LLMs with elevated confidence, regardless of the dataset or prompting method utilized.

\section{Related Work}

The introduction of CoT prompting by ~\citep{wei2022chain} has been a seminal development in highlighting the multi-step reasoning capabilities of LLMs. This approach has paved the way for various refinements and enhancements ~\citep{ zhou2022large,wang2022self,Fu2022ComplexityBasedPF, Zelikman2022STaRBR,Yao2023TreeOT,Besta2023GraphOT}, to enhance the effectiveness of LLMs in addressing complex problems.
For further improvement of the result of CoT, some methods like \citet{wang2022self} and ~\citet{Yao2023TreeOT} concentrate on the optimization of reasoning structures and pathways.While others focused on careful planning and verification of the reasoning process.For example,  ~\citep{Lightman2023LetsVS} introduced the PRM800K dataset, which is notable for its inclusion of step-wise correctness labels obtained through crowdsourcing.And ~\citep{Miao2023SelfCheckUL} extracted insights from multiple reasoning steps and conducted sequential verifications to rectify inconsistencies.

While these works have demonstrated significant advancements in handling complex problems, they often require significantly increased resource. Balancing resource expenditure and efficiency is essential. This research aims to streamline budget allocation and resource utilization while maintaining high efficacy and efficiency in LLM reasoning tasks. The goal is to achieve a harmonious balance between resource deployment and optimal performance, allowing for an efficient and resourceful approach to LLM reasoning.

\section{Conclusion}

In this paper, we introduced a  dynamic decision-making framework called DynaThink, which enables LLMs to make smarter choices between fast and slow problem-solving methods, striking a balance between efficiency and effectiveness. Across various reasoning tasks, DynaThink outperforms existing self-consistency methods, enhancing accuracy while optimizing computational resources.


\section{Limitations}
DynaThink has limitations. The dichotomy of 'Fast' and 'Slow' thinking may oversimplify the complexity of problems, potentially overlooking tasks of intermediate difficulty. This recognition motivates our future work, focusing on refining task categorization methods to enable a more nuanced and adaptable approach. This will empower LLMs to handle a wider range of problem complexities with greater precision and adaptability. The insights gained from this research will drive advancements in the field, leading to the development of LLMs better equipped to handle diverse reasoning challenges.

\bibliography{custom}

\appendix

\section{Algorithm}
\label{sec；algorithm}
The algorithm presents how Dynathink chooses the threshold and recursively separates the whole question set into to subsets.
\begin{algorithm}[ht]
\small
\caption{DynaThink}
\label{alg:algorithm}
\begin{algorithmic}[1]
\Require Problem set, \(P\)
\Ensure Fast thinking question set, \(Q_f\) and Slow thinking question set, \(Q_s\)
\State \(Q_f \gets \varnothing\), \(Q_s \gets \varnothing\) \Comment{Initialize set for fast questions and slow questions}
\State \(n \gets 2\) \Comment{n can be initialized by any integer less than total generation times.}
\Repeat
    \State Generate \(n\) responses for problem set \(P\) by querying the LLM; store as \(Q\).
    \State Initialize question set \(Q_1\), \(Q_2\) and \(Q_3\) as \(\varnothing\).
    \State Calculate voting distribution \(F\) for each question in \(Q\) based on consistency.
    \For{each question \(i\) in \(Q\)}
        \State Determine the answer with the highest votes, \(a_j\), and its vote count, \(max(F(i))\).
        \If{\(max(F(i)) \geq \lfloor \frac{n}{2} \rfloor + 1\)}
            \State Add \(Q(i)\) to \(Q_1\) \Comment{First selected question set}
        \Else
            \State Add \(Q(i)\) to \(Q_2\) \Comment{Set for slow questions}
        \EndIf
    \EndFor
    \For{each question in \(Q_1\)}
        \State Extract the minimum step array from each answer and determine the step distribution.
        \For{each question \(i\)}
            \State Find the minimum steps, \(min(Steps(i))\), and the step of the majority-voted answer, \(a_i\).
            \If{\(a_i == min(Steps(i))\)}
                \State Add \(Q_1(i)\) to \(Q_3\)
            \Else
                \State Add \(Q_1(i)\) to \(Q_2\)
            \EndIf
        \EndFor
    \EndFor
    \If{\(Q_3 \neq \varnothing\)}
        \State Update \(Q_f\) with \(Q_f \cup Q_3\)
        \State Update \(P\) with the questions in \(Q_2\)
        \State increase \(n\) \Comment{n can increase by any integer based on the budget limit}
    \Else
        \State \(Q_s=Q_2\)
    \EndIf
\Until{\(Q_3 = \varnothing\)} \Comment{Continue until \(Q_3\) is  \(\varnothing\)}
    
\Return \(Q_f,Q_s\)
\end{algorithmic}
\end{algorithm}

\section{Prompt Template}
\label{sec:prompt-template}
To let the large language models solve problems in steps, we use the following 
prompt as shown in Table~\ref{tab:problem_solve_prompt}.

\begin{table}[!htbp]
\small
    \centering
    \colorbox{gray!8}{
    \begin{tabular}{@{}p{7.3cm}}
    ============= \textsc{Prompt Example} =============\\\\
    Solve the following problem step by step. Please start each step with "Step :"\\\\

    $[$The problem description$]$ \\\\

    $[$Answer$]$\\
    Step : ......\\
    Step : ......\\
    ... \\
    ... \\
    ...
    \end{tabular}}\caption{Instruction template to prompt ChatGPT to solve the problems in steps.}.
    \label{tab:problem_solve_prompt}
\end{table}

 \renewcommand{\dblfloatpagefraction}{.9}

\begin{figure*}[h]
    \centering
    
    \begin{subfigure}{0.3\textwidth}
        \centering
        \includegraphics[width=\linewidth]{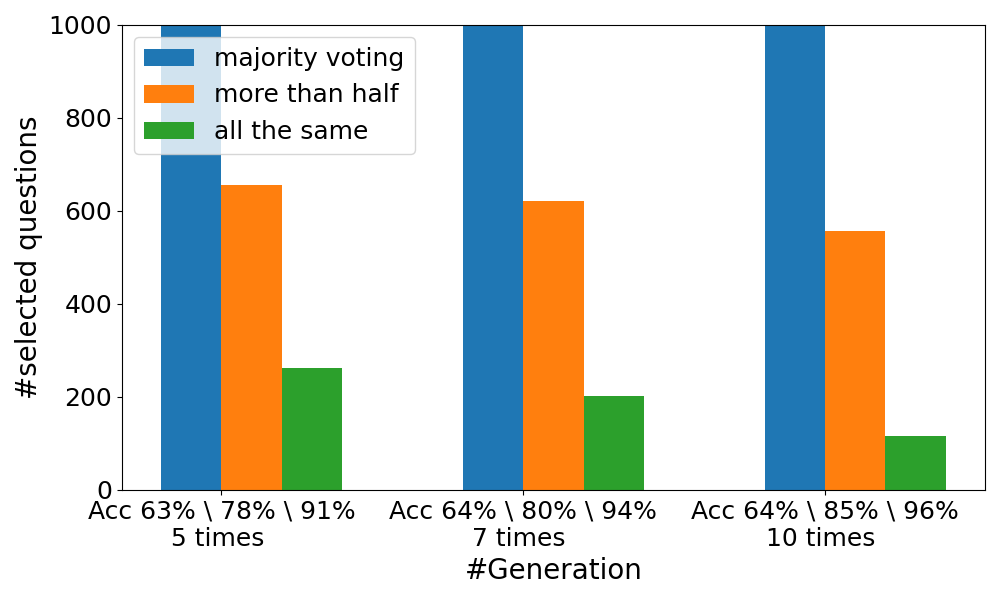}
        \caption{AQuA(zero shot)}
        \label{fig:sub1}
    \end{subfigure}
    \hfill
    \begin{subfigure}{0.3\textwidth}
        \centering
        \includegraphics[width=\linewidth]{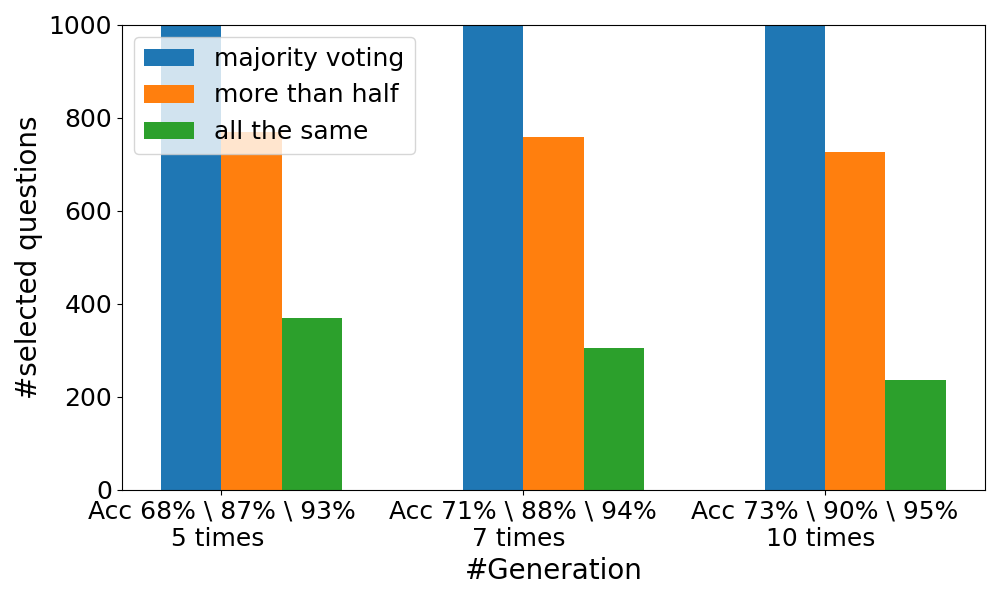}
        \caption{MathQA(zero shot)}
        \label{fig:sub2}
    \end{subfigure}
    \hfill
    \begin{subfigure}{0.3\textwidth}
        \centering
        \includegraphics[width=\linewidth]{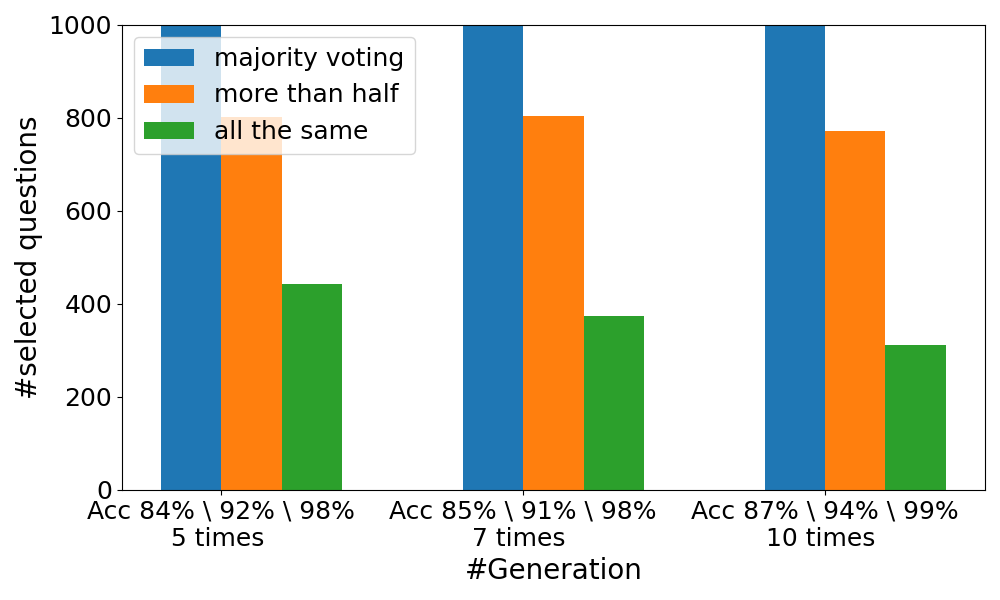}
        \caption{SVAMP(zero shot)}
        \label{fig:sub1}
    \end{subfigure}
    \begin{subfigure}{0.3\textwidth}
        \centering
        \includegraphics[width=\linewidth]{Figures/threshold_selection/zero_shot5/AQuA.png}
        \caption{AQuA(few shot)}
        \label{fig:sub1}
    \end{subfigure}
    \hfill
    \begin{subfigure}{0.3\textwidth}
        \centering
        \includegraphics[width=\linewidth]{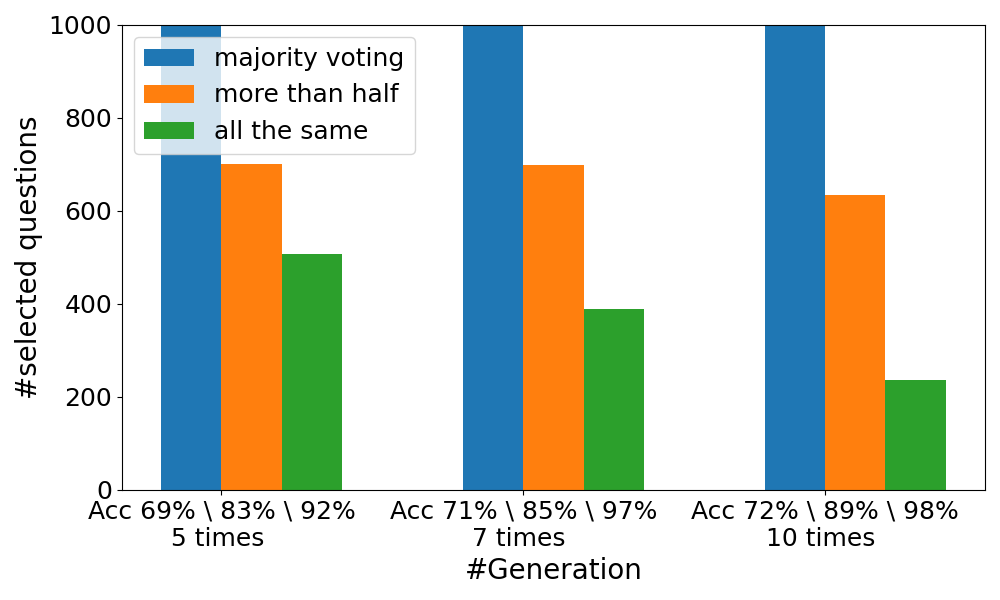}
        \caption{mathqa(few shot)}
        \label{fig:sub2}
    \end{subfigure}
    \hfill
    \begin{subfigure}{0.3\textwidth}
        \centering
        \includegraphics[width=\linewidth]{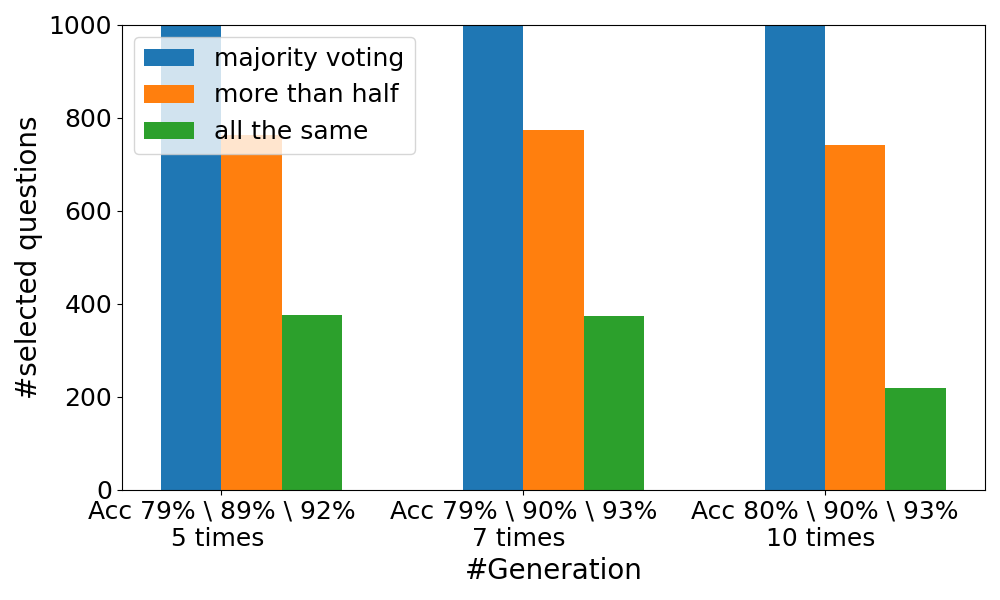}
        \caption{GSM8K(few shot)}
        \label{fig:sub1}
    \end{subfigure}

    \caption{Ablation Study of Consistency Verification. We employed the zero-shot and few-shot COT prompting techniques, to query the GPT-3.5-Turbo  five, seven and  ten times for each question.  Three thresholds for consistency verification in DynaThink are considered, i.e., 
     majority voting, more than half  and all the same.}
    \label{fig:Consistency}
\end{figure*}

\begin{figure*}[h]
    \centering
    
    \begin{subfigure}{0.3\textwidth}
        \centering
        \includegraphics[width=\linewidth]{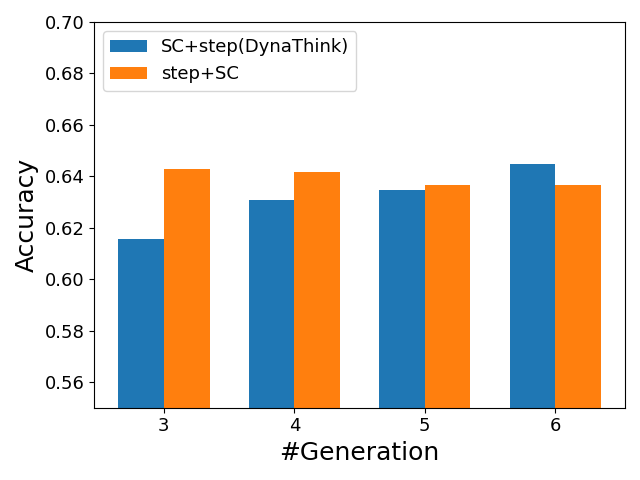}
        \caption{AQuA(zero shot)}
        \label{fig:sub1}
    \end{subfigure}
    \hfill
    \begin{subfigure}{0.3\textwidth}
        \centering
        \includegraphics[width=\linewidth]{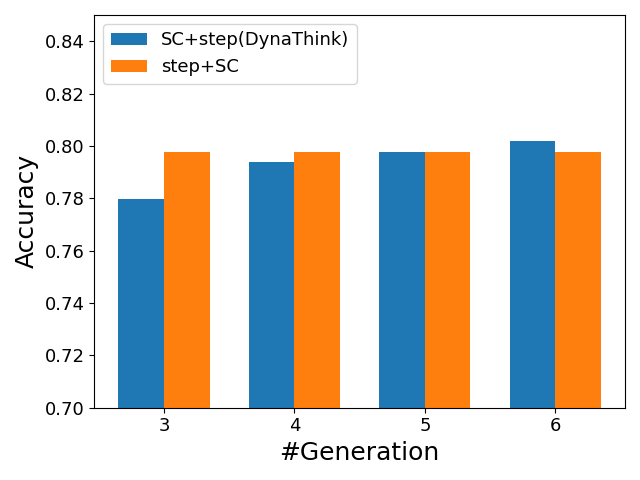}
        \caption{GSM8K(zero shot)}
        \label{fig:sub2}
    \end{subfigure}
    \hfill
    \begin{subfigure}{0.3\textwidth}
        \centering
        \includegraphics[width=\linewidth]{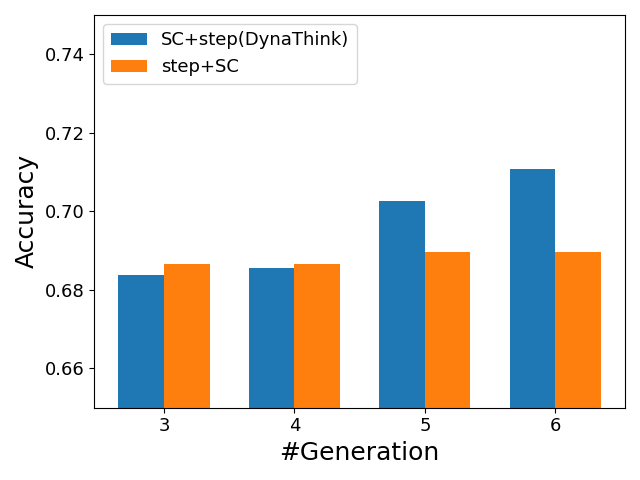}
        \caption{MathQA(zero shot)}
        \label{fig:sub1}
    \end{subfigure}


    \caption{Ablation Study of Verification Order. SC + Step (DynaThink): initially deployed with consistency verification, followed by reasoning complexity verification. Step + SC: initially deployed with reasoning complexity verification, followed by consistency verification. The zero-shot COT prompting technique is utilized to query the GPT-3.5-Turbo six times. }
    \label{fig:verification-order}
\end{figure*}
\section{Ablation Study of Consistency Verification}
\label{sec:Ablation Study of Consistency Verification}

In the development of our DynaThink framework, we have chosen to adopt the "over half votes" criterion as our threshold within the Consistency Verification process. This decision is elucidated by examining the outcomes associated with three distinct voting thresholds: "majority voting," "more than half," and "all the same." These thresholds were evaluated across various datasets, including MathQA, AQuA, and SVAMP, utilizing the COT prompting technique to query the GPT-3.5-Turbo model five, seven, and ten times for each question.

The empirical results, as depicted in Figure~\ref{fig:Consistency}, reveal that the "all the same" voting threshold does indeed enhance the accuracy for the questions selected under this criterion. Nevertheless, this method results in the exclusion of a significant portion of potential questions. On the other hand, the "more than half" voting threshold allows for the inclusion of approximately 80\% of the questions, with only a 4\%-6\% decrease in accuracy compared to the "all the same" voting threshold, which accommodates about 30\% of all questions. This trend holds true across both the AQuA and SVAMP datasets.
\renewcommand{\dblfloatpagefraction}{.9}
\begin{figure*}[h]
    \centering
    \begin{subfigure}{0.3\textwidth}
        \centering
        \includegraphics[width=\linewidth]{Figures/Steps/total/zero_shot4/AQuA.png}
        \caption{AQuA(zero shot)}
        \label{fig:sub1}
    \end{subfigure}
    \hfill
    \begin{subfigure}{0.3\textwidth}
        \centering
        \includegraphics[width=\linewidth]{Figures/Steps/total/zero_shot4/gsm8k.png}
        \caption{GSM8K(zero shot)}
        \label{fig:sub2}
    \end{subfigure}
    \hfill
    \begin{subfigure}{0.3\textwidth}
        \centering
        \includegraphics[width=\linewidth]{Figures/Steps/total/zero_shot4/mathqa.png}
        \caption{MathQA(zero shot)}
        \label{fig:sub1}
    \end{subfigure}

    \begin{subfigure}{0.3\textwidth}
        \centering
        \includegraphics[width=\linewidth]{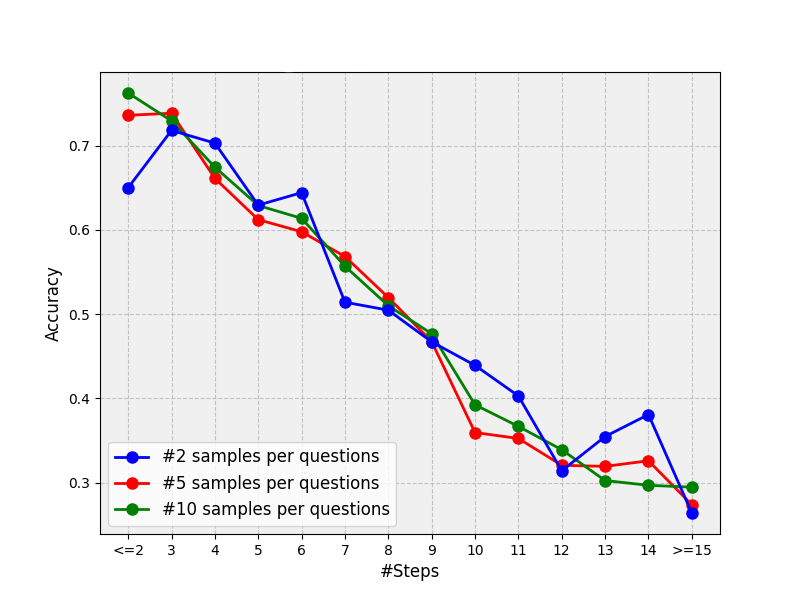}
        \caption{AQuA(few shot)}
        \label{fig:sub1}
    \end{subfigure}
    \hfill
    \begin{subfigure}{0.3\textwidth}
        \centering
        \includegraphics[width=\linewidth]{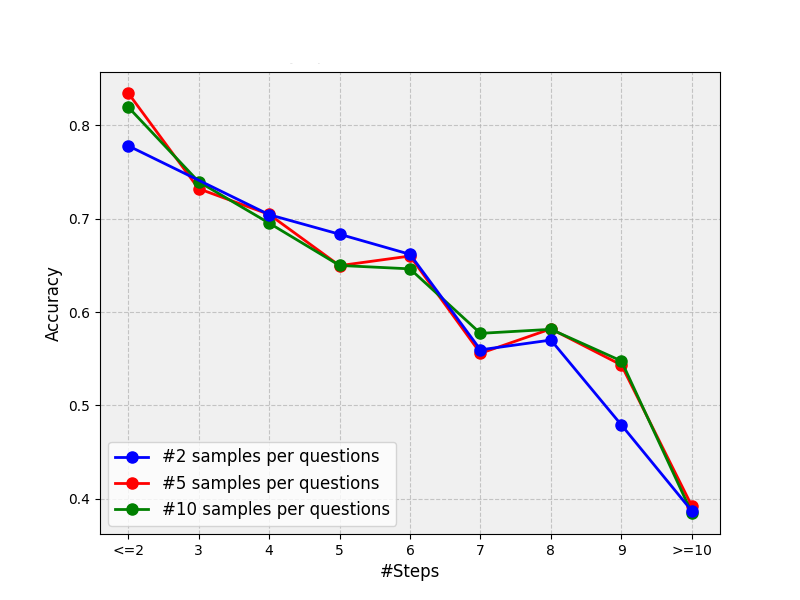}
        \caption{GSM8K(few shot)}
        \label{fig:sub2}
    \end{subfigure}
    \hfill
    \begin{subfigure}{0.3\textwidth}
        \centering
        \includegraphics[width=\linewidth]{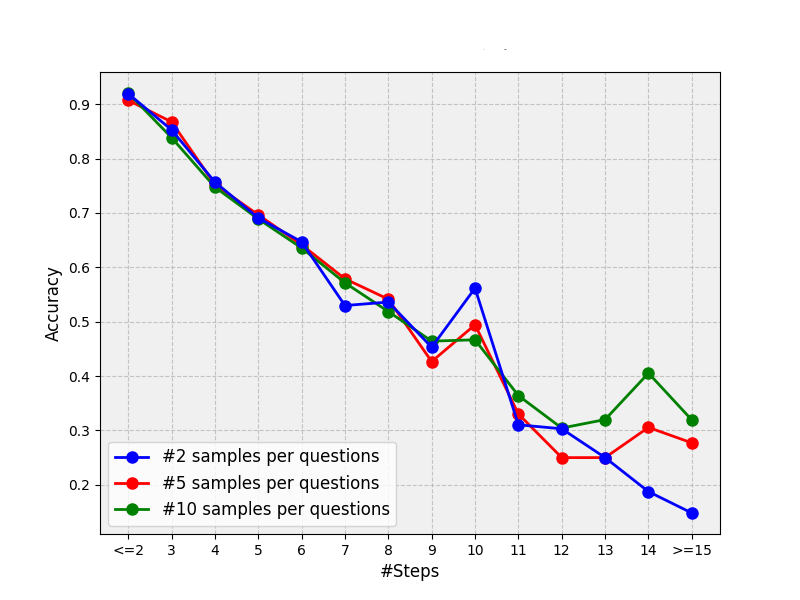}
        \caption{MathQA(few shot)}
        \label{fig:sub1}
    \end{subfigure}

    \caption{The rest of data in ablation study of reasoning complexity verification.}
    \label{fig:rest of ablation study of reasoning complexity verification}
\end{figure*}
These findings suggest that while uniformity in responses can increase the accuracy of the chosen question set, it simultaneously limits the overall number of questions that can be considered. By implementing the "over half votes" criterion, it is possible to secure a subset of questions with relatively high accuracy while also including a larger portion of the questions, thus achieving an optimal balance between quality and quantity.

\section{Ablation Study of Verification Order}
\label{sec:Ablation Study of Verification Order}
This section conducts a thorough analysis of the viability of reversing the sequence of these verifications by inverting the order in which they are conducted. The study is assessed using MathQA, AQuA, and GSM2K test set. The zero-shot COT prompting techniques have been utilized to query the GPT-3.5-Turbo for each question multiple times.
The implications of prioritizing step selection before assessing consistency are illustrated in Figure~\ref{fig:verification-order}.  It can be observed that initiating with reasoning complexity verification may yield higher initial results, yet the overall accuracy tends to stabilize, and in some cases it even experiences a decrease. Conversely, the employment of DynaThink fosters a stable enhancement in results.

\section{Number of test data}
\begin{table}[!htp]
    \centering
    
    \begin{tabular}{l|c}
        \hline
        Dataset  & Total of data\\
        \hline
        MATH & 700\\
        MathQA & 1000\\
        SVAMP & 1000\\
        GSM8K & 1000\\
        AQuA & 1000\\
        StrategyQA & 1000\\
        \hline
    \end{tabular}
    \caption{Number of test data} \label{Number of test data}
 \end{table}

\section{The rest data of ablation study of reasoning complexity verification}
\label{sec:rest data of ablation study of reasoning complexity verification}

All the results of few-shot and part of the results of zero-shot are shown in  Figure~\ref{fig:rest of ablation study of reasoning complexity verification}.

\section{The rest data of main results}
\label{sec:The rest results of main result}

\begin{figure*}[ht]
    \centering
    \begin{subfigure}{0.3\textwidth}
        \centering
        \includegraphics[width=\linewidth]{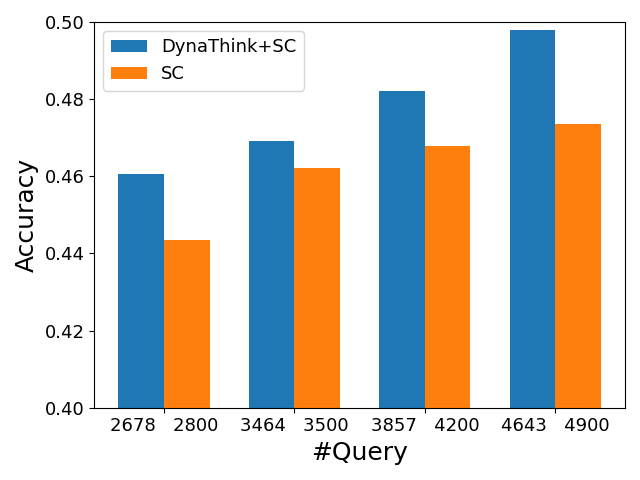}
        \caption{MATH(few shot-GPT-3.5-turbo)}
        \label{fig:sub2}
    \end{subfigure}
    \hfill
    \begin{subfigure}{0.3\textwidth}
        \centering
        \includegraphics[width=\linewidth]{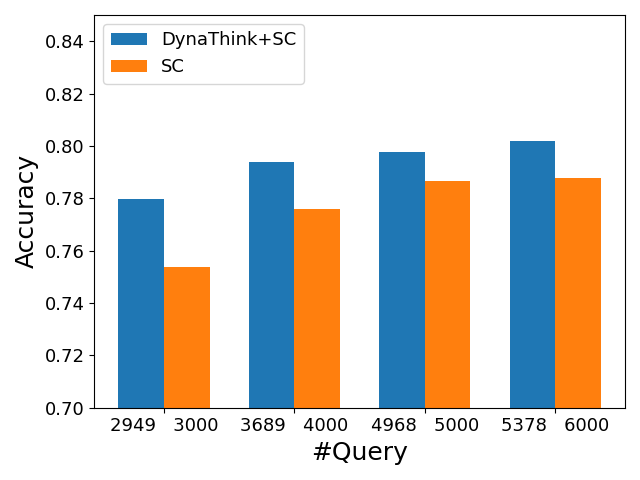}
        \caption{GSM8K(few shot-GPT-3.5-turbo)}
        \label{fig:sub2}
    \end{subfigure}
    \hfill
    \begin{subfigure}{0.3\textwidth}
        \centering
        \includegraphics[width=\linewidth]{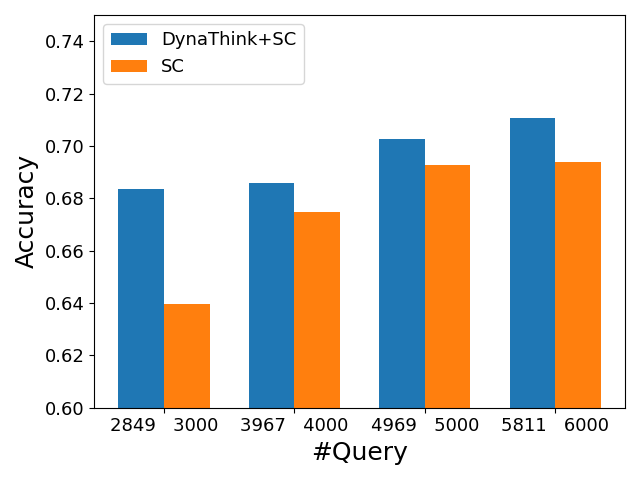}
        \caption{MathQA(few shot-GPT-3.5-turbo)}
        \label{fig:sub1}
    \end{subfigure}
    \begin{subfigure}{0.3\textwidth}
        \centering
        \includegraphics[width=\linewidth]{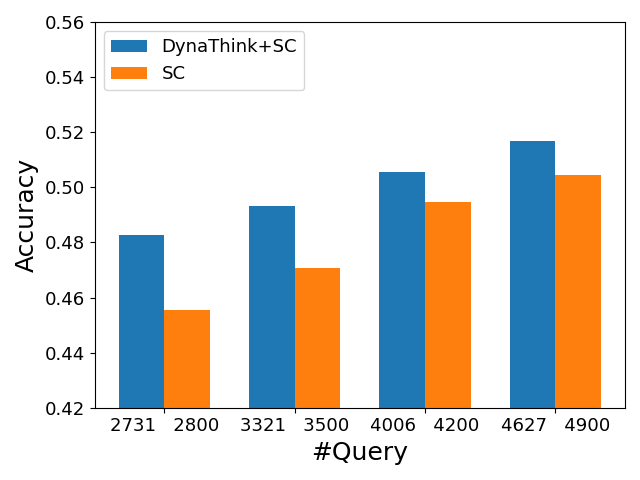}
        \caption{MATH(zero-shot-Gemini)}
        \label{fig:sub2}
    \end{subfigure}
    \hfill
    \begin{subfigure}{0.3\textwidth}
        \centering
        \includegraphics[width=\linewidth]{Figures/main_result/zero_shot/gemini_gsm8k.png}
        \caption{GSM8K(zero-shot-Gemini)}
        \label{fig:sub2}
    \end{subfigure}
    \hfill
    \begin{subfigure}{0.3\textwidth}
        \centering
        \includegraphics[width=\linewidth]{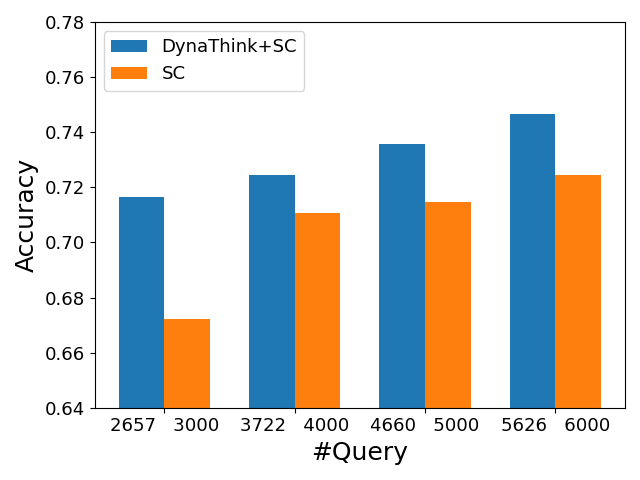}
        \caption{MathQA(zero-shot-Gemini)}
        \label{fig:sub1}
    \end{subfigure}
    \begin{subfigure}{0.3\textwidth}
        \centering
        \includegraphics[width=\linewidth]{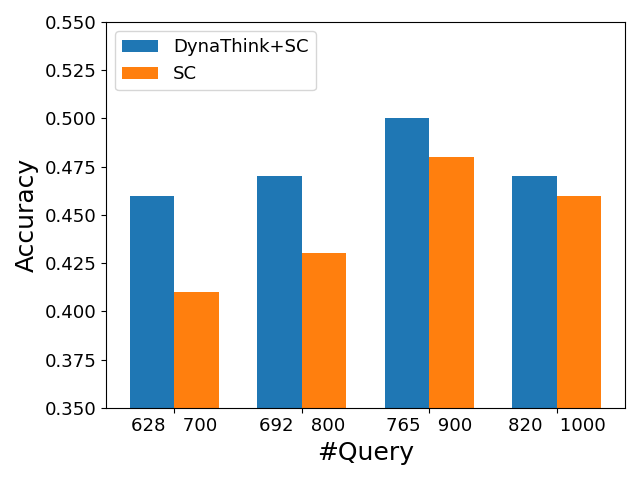}
        \caption{AQuA(Mixtral-8x7B-v0.1)}
        \label{fig:sub2}
    \end{subfigure}
    \hfill
    \begin{subfigure}{0.3\textwidth}
        \centering
        \includegraphics[width=\linewidth]{Figures/main_result/open_source/mistral_gsm8k.png}
        \caption{GSM8K(Mixtral-8x7B-v0.1)}
        \label{fig:sub2}
    \end{subfigure}
    \hfill
    \begin{subfigure}{0.3\textwidth}
        \centering
        \includegraphics[width=\linewidth]{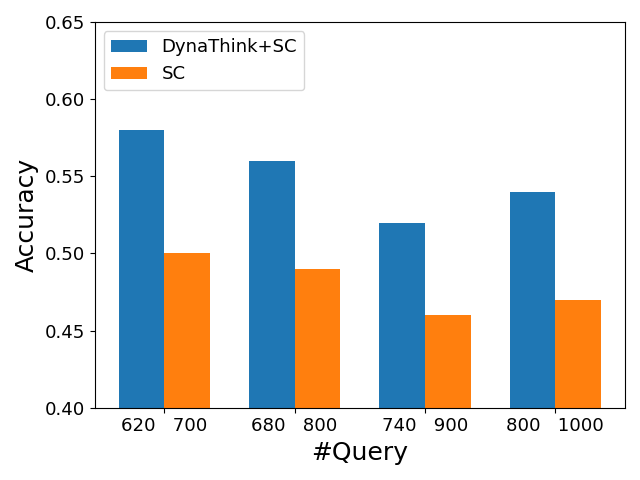}
        \caption{MathQA(Mixtral-8x7B-v0.1)}
        \label{fig:sub1}
    \end{subfigure}
    \begin{subfigure}{0.3\textwidth}
        \centering
        \includegraphics[width=\linewidth]{Figures/main_result/few_shot/math.png}
        \caption{MATH(few shot-GPT-3.5-turbo)}
        \label{fig:sub2}
    \end{subfigure}
    \hfill
    \begin{subfigure}{0.3\textwidth}
        \centering
        \includegraphics[width=\linewidth]{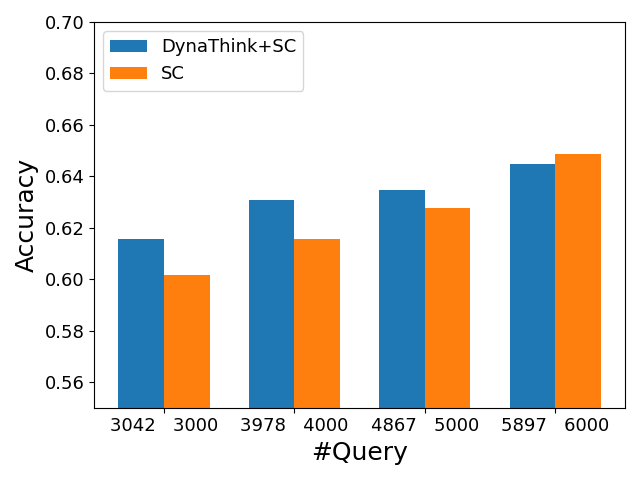}
        \caption{AQuA(few shot-GPT-3.5-turbo)}
        \label{fig:sub1}
    \end{subfigure}
    \hfill
    \begin{subfigure}{0.3\textwidth}
        \centering
        \includegraphics[width=\linewidth]{Figures/main_result/few_shot/gsm8k.png}
        \caption{GSM8K(few shot-GPT-3.5-turbo)}
        \label{fig:sub2}
    \end{subfigure}
    \caption{The rest of data in main results.}
    \label{fig:rest-main-result}
\end{figure*}

All the results of few-shot and part of the results of zero-shot are shown in  Figure~\ref{fig:rest-main-result}.

\section{Generalization to SelfCheck}
\label{sec:Generalization to SelfCheck}
We have  demonstrated the superiority of the DynaThink framework when compared to the  self-consistency baseline. In this section, we  integrate DynaThink with SelfCheck\footnote{\url{https://github.com/NingMiao/SelfCheck}} ~\citep{Miao2023SelfCheckUL}, a recent method that presents a more complex strategy beyond the linearly structured CoT approach, by incorporating careful self-evaluations. We evaluate this enhanced method, which we have termed DynaThink+SelfCheck,  using the MathQA dataset. The experimental setting was designed in alignment with the zero-shot setting, as detailed in the work of ~\citep{Miao2023SelfCheckUL}

Our observations from this experiment reveal that DynaThink+SelfCheck not only maintains a commendable accuracy level comparable to that of the original SelfCheck method—preserving a 72\% accuracy rate—but it also achieves a significant efficiency gain by reducing the number of GPT-3.5-Turbo queries to merely one-quarter of those utilized by the original SelfCheck approach. This finding underscores the effectiveness of DynaThink+SelfCheck in enhancing the operational efficiency of solving complex problems while retaining high accuracy,showcasing its potential as a formidable tool in the domain of advanced reasoning and problem-solving.
\renewcommand\floatpagefraction{.9}
\end{document}